\documentclass[copyright,creativecommons]{eptcs}
\providecommand{\event}{FMAS 2026}

\usepackage{iftex}
\ifpdf
  \usepackage{underscore}
  \usepackage[T1]{fontenc}
\else
  \usepackage{breakurl}
\fi

\usepackage[utf8]{inputenc}  
\usepackage{graphicx}        
\usepackage{amsmath,amssymb} 
\usepackage{orcidlink}
\usepackage{xspace}             
\usepackage{ifthen}

\usepackage{amsmath,amsthm}

\DeclareMathOperator*{\maxintersect}{M}

\newtheorem{theorem}{Theorem}
\newtheorem{lemma}{Lemma}

\theoremstyle{definition}
\newtheorem{definition}{Definition}
\theoremstyle{remark}

\newcommand{\reals}{\mathbb{R}} \newcommand{\nnrats}{\mathbb{Q}}
\newcommand{\initLoc}{l_0} 
\newcommand{\Zones}{\mathit{Zones}} \newcommand{\Act}{\mathit{Act}}
\newcommand{\enab}{g}
\newcommand{\inv}{\mathit{inv}} \newcommand{\prob}{\mathit{prob}}
\newcommand{\reset}{\mathit{reset}} \newcommand{\constraints}[1][{\cal
X}]{\Phi_{#1}} \newcommand{\clocks}{{\cal X}}
\newcommand{\resets}{\mathit{resets}}
\newcommand{\post}[1][(a,e)]{\mathsf{post}#1}
\newcommand{\postt}[1][(a,e)]{\mathsf{post_T}#1}
\newcommand{\postj}[1][(a,e)]{\mathsf{post_J}#1}
\newcommand{\pre}[1][(a,e)]{\mathsf{pre}#1}
\newcommand{\pret}[1][(a,e)]{\mathsf{pre_T}#1}
\newcommand{\prej}[1][(a,e)]{\mathsf{pre_J}#1}
\newcommand{\edges}{\operatorname{outcomes}}
\newcommand{\outcomes}{\edges}

\newcommand{\SYMBA}{\texttt{SYMAR}\xspace}
\newcommand{\sn}{\SYMBA}

\newcounter{algline}
\newcommand{\aln}{\stepcounter{algline}\makebox[1.5em][r]{\footnotesize\thealgline:}}
\newcommand{\alnl}[1]{\refstepcounter{algline}\label{#1}\makebox[1.5em][r]{\footnotesize\thealgline:}}

\newboolean{draft}
\setboolean{draft}{false}

\newcommand{\from}[2]{\ifthenelse{\boolean{draft}}{\textcolor{red}{\noindent\textbf{//}\textbf{Note
from #1:}\ #2\textbf{//}}}{}}

\begin{document}

\title{Symbolic Abstraction Refinement}

\newcommand{\titlerunning}{Symbolic Abstraction Refinement}
\newcommand{\authorrunning}{Bryce, Goldman, Musliner, Gottlieb \& Skovbekk}

\author{Daniel Bryce \orcidlink{0000-0001-5809-3658}
\institute{Smart Information Flow Technologies, LLC\\
Minneapolis MN 55401, USA}
\email{dbryce@sift.net}
\and
Robert P. Goldman \orcidlink{0000-0002-5851-4851}
\institute{Smart Information Flow Technologies, LLC\\
Minneapolis MN 55401, USA}
\email{rpgoldman@sift.net}
\and
David J. Musliner \orcidlink{0009-0004-6022-103X}
\institute{Smart Information Flow Technologies, LLC\\
Minneapolis MN 55401, USA}
\email{musliner@sift.net}
\and
Jeremy Gottlieb \orcidlink{0009-0002-5272-7175}
\institute{Smart Information Flow Technologies, LLC\\
Minneapolis MN 55401, USA}
\email{jgottlieb@sift.net}
\and
John Skovbekk \orcidlink{0000-0002-0470-9055}
\institute{Smart Information Flow Technologies, LLC\\
Minneapolis MN 55401, USA}
\email{jskovbekk@sift.net}
}

\maketitle

\begin{abstract}
Autonomous systems choose actions whose outcomes are uncertain and are subject to
time constraints.  Probabilistic timed
automata (PTAs) model exactly this, and verifying one answers a question about the
system's decision-making: the maximum probability of reaching a goal, or the minimum
probability of reaching a hazard, that a controller can guarantee.  Verification
normally works over symbolic states, each pairing a discrete
location with a region of clock values, and its difficulty is that the best behavior
generally varies \emph{within} such a region.  The PRISM verifier's most capable PTA method
responds by splitting regions until each is uniform, rebuilding the transitions that
reach them and propagating the split backward through the state graph of a stochastic
game, a cascade of graph reconstruction that partitions regions the property never
depends on.  We present \sn, which instead makes the value function symbolic: each
symbolic state carries a function from sub-regions of its clock region to probability
bounds, so refinement is the appearance of new sub-regions in that function, and
previously generated states are never split nor their incoming transitions
reconstructed.  Both sources of PTA nondeterminism, the timing and the choice of
actions, are captured implicitly by the preimage computation that produces those
sub-regions.  Heuristic forward search then bounds how much refinement is done at
all, expanding an And-Or graph from the initial state along the most promising
behaviors so that effort is confined to the property-relevant fragment, guided by
bounds and preferred actions from a single unrefined run of the existing method; a
partial-refinement backup keeps the otherwise combinatorial sub-region count lean, and every stored interval provably contains the true probability.  On the
PRISM PTA benchmarks, whose subproblems recur in autonomous systems (distributed
agreement under contention, leader election, and self-configuration without
central coordination), \sn is up
to $108\times$ faster, solves an instance on which the existing method exhausts the
time limit, and is markedly less sensitive to the deadline constant, resolving one
instance in a $37$-node search graph, unchanged as the deadline triples, where the
existing method needs $86{,}380$ abstract states and $73$ refinement rounds.
\end{abstract}


\section{Introduction}

Autonomous systems by their very nature make decisions about what actions to take and
when to take them, and they can sometimes fail.  A critical aspect of designing these
systems is ensuring they can avoid foreseeable failures.  Probabilistic timed automata
(PTAs) are a natural fit for these concerns.  PTAs provide a powerful formalism for
modeling real-time systems with uncertainty, finding application in autonomous systems
\cite{Mu2023Runtime}, wireless protocols
\cite{Kwiatkowska2002IEEE80211,Kwiatkowska2006DigitalClocks}, security protocols
\cite{Lanotte2005NonRepudiation,SiedleckaLamch2021}, and embedded control systems
\cite{Mutsuda2005EmbeddedSystems}.  More broadly, probabilistic
model checking is applied throughout autonomy (synthesising mission controllers for
unmanned aerial vehicles \cite{Hoffmann2016AutonomousAgent,Giaquinta2018Strategy},
verifying human--robot handover tasks \cite{Webster2020Corroborative}, and assuring
robots whose perception components are learned \cite{Ye2023Agriculture}) with
deployments in mobile service robotics and underwater vehicles
\cite{Kwiatkowska2025Trends}.  Such systems act under deadlines, so the timing of
their decisions is part of what must be verified.

A PTA defines an MDP over
symbolic states, pairs of a location and a clock zone, and verification amounts to
computing the optimal reachability probability over that symbolic state space.
A PTA separates two kinds of choice: a \emph{controller} selects which action to take,
and a \emph{scheduler} determines when actions occur.  The actions have stochastic
outcomes and event timing can be partially controlled or even adversarial.  The optimal reachability probability is then a guarantee about
that decision-making: the maximum probability of reaching a goal any controller can
achieve, or the minimum probability of reaching a hazard any controller can be held
to, against a worst-case environment.  Verifying the decisions themselves rather than
a nondeterministic over-approximation requires resolving both choices separately,
which is what makes PTA verification harder than verification of the underlying
untimed MDP.  The benchmarks of Section~\ref{sec:results} exercise exactly the
capabilities an autonomous system depends on: agreeing on a leader, resolving
contention for a shared channel, configuring an address without central coordination,
and completing an exchange in the presence of an adversary, each under a deadline,
and each a case where the answer is a probability rather than a yes or no.

Existing methods compute a \emph{scalar} value function.  Whether the symbolic state
space is treated as an MDP and solved by value iteration
\cite{kwiatkowska2002automatic,kwiatkowska2007symbolic} or as a stochastic game in
which one player chooses a concrete state within the zone and the other chooses among
the actions then enabled \cite{KNP09c}, each symbolic
state is assigned a single value: one probability bound, or one pair of bounds,
per (location, zone) pair.  Since the optimal behavior generally varies
\emph{within} a zone, such a value is sound only once the zone is fine enough that
one behavior is optimal throughout it.  Achieving that means splitting the state
into finer states and rebuilding the transitions that reach it, and because each
predecessor was solved against the coarser partition, the split propagates backward
through the reachable state space.  This global refinement cascade is the
scalability bottleneck: it is driven by the representation's inability to express
variation within a zone, not by the property, so effort is spent partitioning
regions the property never depends on.

\sn (\textbf{SYM}bolic \textbf{A}bstraction \textbf{R}efinement) keeps the
abstraction-refinement insight of the stochastic games (SG) method
\cite{KNP09c}, that a symbolic state admitting several distinct optimal
behaviors must be refined before its bounds converge, while changing two things
about how refinement is carried out.  First, it is performed \emph{symbolically}, in
the value function.  The central object is $V : \Zones \rightarrow [0,1]^2$, a
\emph{zone-valued probability function} mapping sub-regions of a state's clock zone
to probability bounds on reaching the target: where a scalar method holds one
interval per state, $V$ holds a partition of that state's zone into cells, each with
its own interval, and refinement is the appearance of new cells.  Both sources of
nondeterminism are handled implicitly by the preimage computation that
defines the value of a transition (Section~\ref{sec:symba}), so the scheduler's zone
distinctions emerge from the geometry of the preimage instead of being materialized
as states, which also removes the need for an explicit two-player game, since a
single node type per symbolic state suffices.  The cascade does not disappear: a
refinement still propagates to predecessors whose bounds came from the coarser value
it replaces.  What changes is its content: in the SG method predecessors are split
and their transitions rebuilt, whereas in \sn the cascade revises values while
leaving existing states and transitions intact.

Second, refinement is \emph{directed by heuristic forward search}.  \sn grows an
And-Or graph forward from the initial state with AO*
\cite{Martelli1978Decision,Nilsson1980Principles}, expanding along the best
partial solution and backpropagating $V$ through ancestors.  Where the SG method
refines globally and uniformly, AO* refines only where the property-relevant
probability mass lies, leaving irrelevant regions at their initial coarse bounds.
The two changes are complementary: symbolic refinement makes each refinement cheap,
and forward search makes most refinements unnecessary.  Two costs come with the
design.  A state's value is computed from a cross product over its successors'
sub-regions, whose size grows combinatorially, so \sn refines each state
\emph{partially}, committing to the sub-regions that currently determine the bounds
and refining further only once they prove to depend on what was deferred.  And
because forward search is only as good as its guidance, \sn takes admissible bounds
and preferred actions from a single unrefined run of the SG method, so a coarse
solution directs the symbolic one.  $P_{\min}$ queries are supported as the
dual of $P_{\max}$ (maxima become minima and upper bounds become lower bounds), with
every extremum-selection site branching on the query direction.

We prove that every interval \sn stores brackets the optimal probability of acting at
each valuation its cell represents, and that the reported root interval contains the
true answer, independently of convergence.  Empirically, on
the PRISM PTA benchmarks \sn is up to $108\times$ faster than the SG method and
solves a csma instance on which that method exhausts the 20 minute time limit; its runtime is
far less sensitive to the deadline constant, resolving one firewire instance in a
$37$-node search graph where the SG method needs $86{,}380$ abstract states over $73$ refinement rounds.

\section{Background}
\label{sec:background}

We define PTAs and the associated reachability problem solved by this work.

\begin{definition}[PTA]
    A PTA is a tuple $\mathsf{P} = (\clocks, L, \initLoc{}, \Act, \inv, \enab,
    \prob)$ where:
    \begin{itemize}
      \item $\clocks$ is a finite set of non-negative real clock variables;
        \item $L$ is a finite set of locations and $\initLoc{}\in L$ is the
              initial location;
        \item $\Act$ is a finite set of actions;
        \item $\inv:L\rightarrow \constraints$ is the invariant condition;
        \item $\enab: L \times \Act \rightarrow \constraints$ is the
          \emph{guard} of an act in a state;
        \item $\prob: L \times \Act\rightarrow {\sf Dist}(2^{(L \times
              2^{\clocks})})$ is the probabilistic transition function, a
              distribution over location-reset pairs, or \emph{outcomes},
              $\outcomes(l, a)$.  Outcomes are ordered, so we may refer to the
              $i^{th}$ outcome of $a$ in $l$, and we write $\resets(l,a,i)$ for
              the clocks it resets.  We use $\outcomes(l, a)$ for both the
              outcomes and their indices.
    \end{itemize}
\end{definition}

A \emph{clock valuation} $v \in \reals^{\cal X}$ assigns each clock a value, and a
state $(l,v) \in L \times \reals^{\cal X}$ pairs a location with a valuation.
Guards and invariants are \emph{clock constraints}, conjunctions and disjunctions
of atoms $x \leq c$, $c \leq x$ and $x + c_1 \leq y + c_2$ with $c \in \nnrats$.
A conjunction of atoms denotes a convex set of valuations called a \emph{zone};
a disjunction denotes a union of zones, which need not be convex.  We write
$\constraints[\clocks]$ for the constraints over $\clocks$, and say a state
satisfies a constraint when its valuation does.
\begin{figure}[tb]
\centering\includegraphics[width=0.4\linewidth]{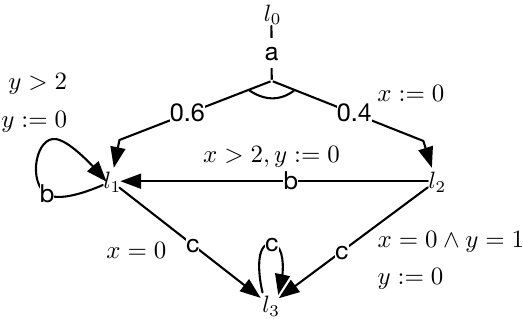}
\caption{\label{fig:ex-pta} Example PTA, adapted from \cite{KNP09c}.  The $a$
transition is probabilistic, leading to $l_1$ and $l_2$.  Taken immediately, $c$
in $l_1$ reaches the target $l_3$; taken after one time unit, $c$ in $l_2$ does.
No execution lets both outcomes of $a$ reach $l_3$.}
\end{figure}

We use the PTA ${\sf P}$ of Figure \ref{fig:ex-pta} as a running example.  It has
$\clocks = \{x,y\}$, $L = \{l_0,\ldots,l_3\}$, $Act = \{a,b,c\}$ and trivial
invariants, with the guards and resets shown in the figure. 

The semantics of a PTA is a probability over its \emph{traces}, strings of two
kinds of transition: time may elapse, or an action may occur.  Only the actions
carry probability, so a trace's probability is the product of the action
probabilities along it.  Two forms of nondeterminism must be resolved before that
product is determined: \emph{when} an action occurs, fixed by the duration of the
preceding time elapse and referred to as \emph{scheduler} nondeterminism, and
\emph{which} enabled action occurs, referred to as \emph{controller}
nondeterminism.  A \emph{policy} resolves both, making a decision at each
nondeterministic choice, and the reachability probability we seek is an optimum
over policies.  A policy still induces a distribution over uncountably many
traces, since time may advance by any real amount, but the probability of reaching
a target depends only on the sequence of action occurrences a trace contains, not
on the exact clock valuations along it.  Traces are therefore grouped into
equivalence classes by those sequences, and the probability of a class is the
product of its action probabilities.  The definitions below make the two
transitions precise and state the problem.

\begin{definition}[Time transition]\label{def:timeTrans} A time transition,
$(l, v_i) \stackrel{n}{\longrightarrow} (l, v_j)$ models the elapse of $n \in
\reals_{\geq 0}$ time units, incrementing every clock by $n$ without changing location:
$v_j[x] = v_i[x] + n$ for every $x \in \clocks$.  For this transition to be
valid, $v_i$ and $v_j$ must both satisfy $\inv(l)$, which entails that every
intermediate valuation does too.
\end{definition}

\begin{definition}[Jump transition]\label{def:jumpTrans} A jump transition,
$(l_i, v_i) \stackrel{a,k}{\longrightarrow} (l_j, v_j)$ models the occurrence of
an action, $a$. For this transition to be valid, $v_i$ must satisfy $\enab(l_i,
a)$, $\prob(l_i, a)[k] > 0$,  $v_j[x] = v_i[x]$ if $x \not\in \resets(l_i, a,
k)$, $v_j[x] = 0$ if $x \in \resets(l_i, a, k)$, and $v_j$ satisfies $\inv(l_j)$.
The probability of this transition, given that $a$ occurs in $v_i$, is the
probability of the $k^{th}$ outcome of $(l_i, a)$, and is independent of the clock
values so long as they satisfy the guard.
\end{definition}

\begin{definition}[Probabilistic Reachability]\label{def:reach}
    The probabilistic reachability problem is to find the maximum (minimum)
    probability, over all policies, of reaching a target $F \subseteq L$ from the
    initial state $\initLoc{}$ of PTA ${\sf P}$, denoted $p_{\sf P}^{max}(F)$
    ($p_{\sf P}^{min}(F)$).  Under a given policy this is the probability of the
    set of trace equivalence classes ending in $F$.
\end{definition}

\section{Symbolic State MDP}
\label{sec:mdp}

A PTA defines an MDP over \emph{symbolic states}.  Each symbolic state
${\sf z} = (l, \zeta)$ is a pair of a location $l$ and a clock zone $\zeta$
(a set of clock valuations expressed as a clock constraint).  The symbolic state
${\sf z}_0 = (l_0, \zeta_0)$, where $\zeta_0 = inv(l_0) \wedge
\nearrow \bigwedge_{x\in {\cal X}} (x=0)$, is the initial symbolic state
representing all clock valuations reachable from $x = 0$ for all $x \in {\cal
X}$ by time transitions.  The operators $\nearrow \zeta$ and
$\swarrow \zeta$ let time in $\zeta$ advance forward or backward.
The operators $\reset(\zeta, X)$ and $unset(\zeta, X)$ set the clocks in $X$ to
zero, or to their possible values before a reset; they are distinct from
$\resets(l,a,i)$, which names the set of clocks that outcome $i$ of $a$ in $l$
resets.  The successor and
predecessor functions below are built from these.

\begin{definition}[Symbolic Successor Function]
  The successor function identifies the states ${\sf z}'$ reached by
  an outcome of $a$ when executed in ${\sf z}= (l,\zeta)$.  The jump step $\postj[]$ restricts to valuations
  meeting the guard and applies the resets; the temporal step $\postt[]$ adds the
  valuations reachable by letting time elapse within the invariant:
\begin{align*}
  \post[]((l, \zeta), a) &= \{\post[]((l, \zeta), a, i) \mid i \in [1, n]\}\\
  \post[]((l, \zeta), a, i) &= \postt[](\postj[]((l, \zeta), a, i))\\
  \postj[]((l, \zeta), a, i) &= (l', \reset(\zeta \wedge \enab(l, a), \resets(l, a, i)))\\
  \postt[]((l, \zeta)) &= (l, \inv(l) \wedge \nearrow\zeta)
\end{align*}
where $n = |\outcomes(l,a)|$ is the number of outcomes of $a$ in $l$, and $l'$
  is a successor location.
\end{definition}

\begin{definition}[Symbolic Predecessor Function]\label{def:pre}
  The predecessor function identifies the valuations of ${\sf z} = (l,\zeta)$ from
  which taking $a$ can reach the given zone in \emph{every} outcome.  Per
  outcome, $\pret[]$ runs time backward within the successor's invariant and
  $\prej[]$ inverts that outcome's resets and imposes the guard and source
  invariant; the results are intersected, since one valuation must serve all $n$
  outcomes of the same occurrence of $a$:
\begin{align*}
\pre[]((l, \zeta), a, \{(l_1, \zeta_1), \ldots, (l_n,\zeta_n)\}) &=  \zeta \wedge \bigwedge\limits_{i=1}^n \prej[](l, a, \pret[]((l_i, \zeta_i)), i)\\
\pret[]((l_i, \zeta_i)) &= \inv(l_i) \wedge \swarrow\zeta_i\\
\prej[](l, a, \zeta', i) &= \enab(l, a) \wedge \inv(l) \wedge unset(\zeta', \resets(l, a, i))
\end{align*}
\end{definition}

The key difficulty in computing optimal reachability probabilities over this MDP
is that in a symbolic state ${\sf z} = (l, \zeta)$, the available actions and
their successors may depend on \emph{which sub-zone of $\zeta$} the clock
valuation lies in.  The guard $\enab(l,a)$ and the reset together determine
which valuations can reach which successor zone.  This means the optimal
probability is not uniform over $\zeta$: it varies by sub-region.  For
example, in the initial symbolic state $(l_0, x=y)$ of the running example,
taking action $a$ immediately (when $0=x=y$) vs.\ after a delay (when $0<x=y$)
leads to different successor zones with different probabilities.

The established response, and the state of the art for PTA verification, is the
stochastic games (SG) method of Kwiatkowska et al.~\cite{KNP09c}; we describe it
only as far as the contrast with \sn requires.  It assigns each symbolic state a
single pair of bounds $[lb, ub]$, obtained from a game in which one player picks a
concrete state inside the zone and the other picks among the actions then enabled:
resolving both cooperatively gives $ub$, resolving the first adversarially gives
$lb$.  The interval is always sound, and its
width indicates the utility in refining it: a wide interval means the state is too abstract.  The remedy is to split ${\sf z}$ into
sub-regions on which the players no longer disagree and rebuilding the
transitions that reach them; iterating split-and-resolve converges to the true
probability.  The cost is that a split is not local: every symbolic state whose
transitions reach ${\sf z}$ was solved against the coarser partition, so each
must be updated in turn, and those updates may force further splits, a
cascade that can reach back to the initial state over many refinement rounds.

\begin{figure}[tb]
\centering\includegraphics[width=\linewidth]{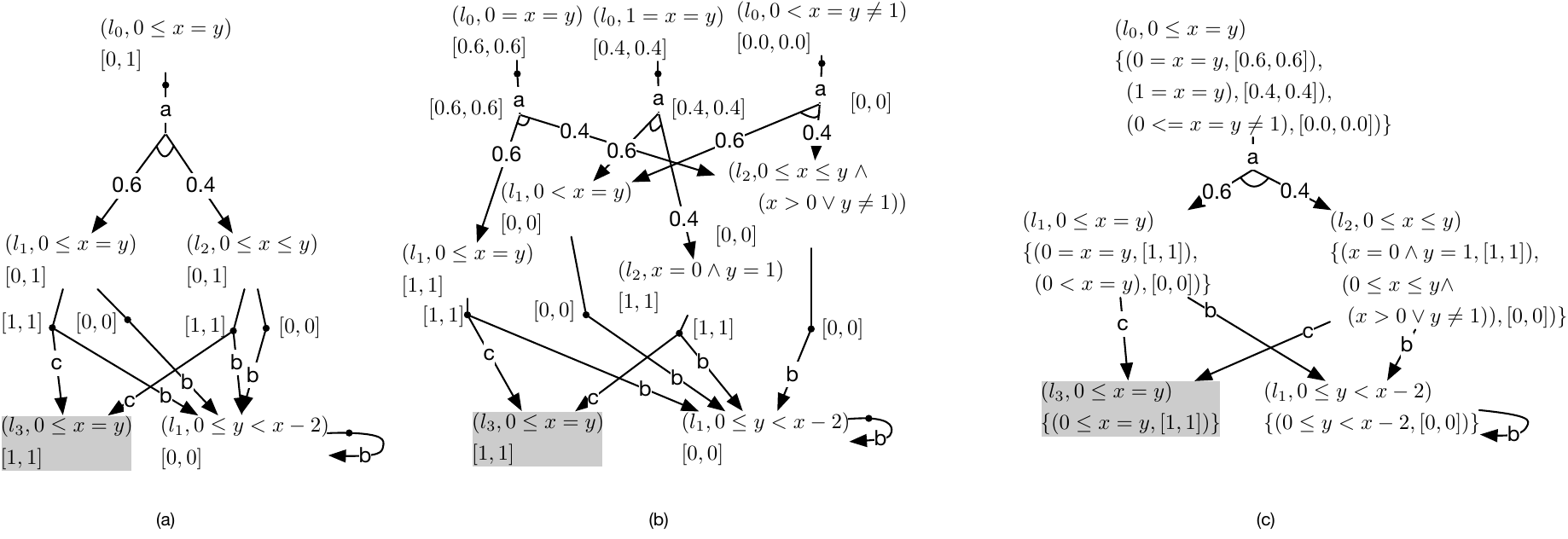}
\caption{\label{fig:ex-pta-sg} Stochastic game and symbolic formulations of the
example PTA.  (a) The initial game.  (b) After refining $(l_1, 0\leq x = y)$ into
$(l_1, 0=x=y)$ and $(l_1, 0<x=y)$, and $(l_2, 0\leq x\leq y)$ into
$(l_2, x=0\wedge y=1)$ and $(l_2, 0\leq x\leq y\wedge(x>0\vee y\neq 1))$.
Symbolic states are labeled with their bounds, e.g.\ $[0.6,0.6]$; the target
state is gray; small black circles denote joint symbolic transitions, with edges
labeled by action and intervals showing the intermediate values from which state
bounds are computed.  (c) \sn's symbolic backpropagation over the same PTA: the
distinctions (b) obtains by splitting states appear instead as cells of the
zone-valued function attached to the unsplit states, so a graph more similar to (a) is
retained throughout.}
\end{figure}

Figure~\ref{fig:ex-pta-sg}(a,b) illustrates the pattern.  In the initial abstraction
(a), the states reached by action $a$ carry uninformative bounds $[0,1]$:
whether $l_1$ or $l_2$ leads to the target depends on when $a$ was taken, a
distinction those states do not make.  Refining $(l_1, 0\leq x=y)$ and
$(l_2, 0\leq x\leq y)$ (b) makes the distinction and the bounds tighten to exact values.  Note
what this costs: the transitions out of ${\sf z}_0$ must be rebuilt to point at
the new sub-states, even though ${\sf z}_0$ itself was never the state in
question.  That obligation is the cascade.

\sn adapts the concept of the SG method, where a symbolic state whose zone
admits several distinct optimal behaviors is too coarse and must be refined.  
Instead of splitting symbolic states, \sn associates a
\emph{zone-valued probability function}
$V_{{\sf z}} : \Zones(\clocks) \rightarrow [0,1]^2$ with each symbolic state
${\sf z}$.  This function maps sub-regions of $\zeta$ to probability bounds
and is computed directly from the preimage computation during backpropagation,
as we describe in the next section.  The partition that the SG method would
materialize as new states appears instead as the cell structure of
$V_{{\sf z}}$.

To be precise about what this eliminates: the refinement still happens and still
propagates, since when $V_{{\sf z}}$ acquires a finer cell structure every
predecessor holds bounds derived from the coarser function and must be
recomputed.  \sn therefore has a cascade too, but it revises values over a graph
whose existing states and transitions are left intact, reusing the preimages
cached on them,
rather than rebuilding the graph the next round of value iteration runs over.
The saving is not the absence of backward propagation but the absence of
reconstruction.  Section~\ref{sec:aostar} then adds the second ingredient,
forward search, which decides \emph{which} symbolic states are worth refining at
all.

\section{Zone-Valued Probability Functions}
\label{sec:symba}

The primary insight of \sn is that the scheduler's zone distinctions, which the
SG method materializes through state splitting, can instead be captured directly
in the value function.  We associate with each symbolic state ${\sf z}$ a
\emph{zone-valued probability function}:

\begin{definition}[Zone-Valued Probability Function]
A zone-valued probability function for symbolic state ${\sf z}$ is a finite set
\[
  V_{{\sf z}} = \{(\zeta_1, [lb_1, ub_1]), \ldots, (\zeta_k, [lb_k, ub_k])\}
\]
where $\{\zeta_i\}$ partition the clock zone of ${\sf z}$.  Each pair
$(\zeta_i, [lb_i, ub_i])$ associates a sub-region of the state zone with
bounds on the probability of reaching the target when an action is taken at a
valuation in that sub-region \emph{now}, before any further delay.  Delay before an
action enters through the preimage (Definition~\ref{def:pre}), which maps each
successor cell back through post-jump elapse, and, at the root only, through time
closure (Section~\ref{sec:alg}).  We call $V_{{\sf z}}$ a
\emph{cell map}: each $(\zeta_i, [lb_i, ub_i])$ is a \emph{cell}.
\end{definition}

Three cases are immediate.  A symbolic state whose location lies in the target
set $F$ has $V_{{\sf z}} = \{(\zeta, [1,1])\}$, and one from which no action is
ever enabled has $V_{{\sf z}} = \{(\zeta, [0,0])\}$; both are exact over the
entire zone.  A symbolic state whose value has not yet been computed carries the
uninformative $V_{{\sf z}} = \{(\zeta, [0,1])\}$, a single cell spanning the
zone.  Every other symbolic state takes its bounds from its successors, via the
two definitions that follow.

\subsection{Transition Value $V_\theta$}

Let $\theta = ({\sf z}, a, \{{\sf z}_1, \ldots, {\sf z}_n\})$ be a symbolic
transition, where ${\sf z}_i$ is the successor for outcome $i$ and carries value
function $V_{{\sf z}_i}$.  Then
\begin{equation}\label{eq:and}
  V_\theta = \maxintersect \Big\{ \big(\pre[]({\sf z}, a, \zeta(c)),\;
    \textstyle\sum_{i=1}^{n} \prob(l,a,i)\cdot c_i \big)
    \;\Big|\; c \in C(\theta),\; \pre[]({\sf z}, a, \zeta(c)) \neq \perp \Big\}
\end{equation}
where $C(\theta) = V_{{\sf z}_1} \times \cdots \times V_{{\sf z}_n}$ is the
cross product of the successors' value functions, $c_i$ is the probability
interval of the cell $c$ selects from $V_{{\sf z}_i}$, and $\zeta(c)$ is the
tuple of zones of those cells.

A \emph{combination} $c \in C(\theta)$
picks one cell from each successor, hypothesizing that outcome $i$ lands in the
sub-region $c$ selects there.  Its \emph{preimage}
$\pre[]({\sf z}, a, \zeta(c))$ is the set of valuations in ${\sf z}$ for which
that hypothesis holds: those where $a$ is enabled \emph{and} every outcome lands
in its selected zone.  Combinations whose preimage is empty ($\perp$) are
discarded, since no valuation can realize them.  Over the valuations that remain,
the outcome intervals are fixed, so the transition's value there is their
\emph{expectation} under $\prob(l,a,\cdot)$, computed by interval arithmetic.
Finally $\maxintersect$ resolves the cells against one another, keeping the
higher $ub$ (then $lb$) wherever two preimages overlap.

Both sources of nondeterminism enter through the preimage.  Distinct
combinations have distinct preimages, so the cross product partitions ${\sf z}$
by which timing and action choice is optimal: the refinement, obtained without
splitting the state.  The cost is that $|C(\theta)|$ is a product over
successors; Section~\ref{sec:aostar} bounds it.
\subsection{Symbolic State Value $V_{{\sf z}}$}

$V_{{\sf z}}$ is the pointwise maximum over applicable transitions: for each
sub-zone $\zeta'$,
\[
  V_{{\sf z}}(\zeta') = \max_{\theta} V_\theta(\zeta')
\]
where the max is by $ub$ then $lb$, merging the zone partitions of all
applicable transitions and taking the best-action value in each sub-region.
A valuation at which no applicable transition is enabled lies in no cell of any
$V_\theta$ and receives $[0,0]$, so $V_{{\sf z}}$ covers the whole zone of ${\sf z}$
even though each $V_\theta$ covers only the region where $a$ is enabled.

\subsection{Running Example}

Figure~\ref{fig:ex-pta-sg}(c) illustrates the \sn symbolic backpropagation for
the running example.  The successors of transition $\theta_a$ (action $a$ from
${\sf z}_0 = (l_0, 0\leq x=y)$) are ${\sf z}_1 = (l_1, 0\leq x=y)$ and
${\sf z}_2 = (l_2, 0\leq x\leq y)$.  Their value functions are
\begin{align*}
V_{{\sf z}_1} &= \{(\zeta_{11},\; [1,1]),\;(\zeta_{12},\; [0,0])\} \\
V_{{\sf z}_2} &= \{(\zeta_{21},\; [1,1]),\; (\zeta_{22},\; [0,0])\}
\end{align*}
for $\zeta_{11} = (0=x=y)$, $\zeta_{12} = (0<x=y)$, $\zeta_{21} = (x=0\wedge
y=1)$ and $\zeta_{22} = (0\leq x\leq y\wedge(x>0\vee y\neq 1))$, the
probabilities for different clock valuations in each successor. 

The combinations $C(\theta_a)$ are all pairs of one element from $V_{{\sf z}_1}$
and one from $V_{{\sf z}_2}$, so the four combinations give
\begin{align*}
 \pre[]({\sf z}_0, a, \{\zeta_{11}, \zeta_{21}\}) &= \perp &&\\
 \pre[]({\sf z}_0, a, \{\zeta_{12}, \zeta_{21}\}) &= (1=x=y) & 0.6[0,0]+0.4[1,1] &= [0.4,0.4]\\
 \pre[]({\sf z}_0, a, \{\zeta_{11}, \zeta_{22}\}) &= (0=x=y) & 0.6[1,1]+0.4[0,0] &= [0.6,0.6]\\
 \pre[]({\sf z}_0, a, \{\zeta_{12}, \zeta_{22}\}) &= (0<x=y\neq 1) & 0.6[0,0]+0.4[0,0] &= [0,0]
\end{align*} 
The first combination is infeasible: reaching $\zeta_{11}$ requires taking $a$ at
$x=y=0$ and reaching $\zeta_{21}$ requires taking it at $x=y=1$, which are
mutually exclusive.  The second and third preimages cover disjoint sub-zones
($1=x=y$ and $0=x=y$), while the fourth covers the remainder $0<x=y\neq 1$.
Together they partition ${\sf z}_0 = (0\leq x=y)$ into three cells, matching the
zone-valued function shown in Figure~\ref{fig:ex-pta-sg}(c).  Since $\theta_a$ is
the only transition applicable in ${\sf z}_0$, $V_{{\sf z}_0} = V_{\theta_a}$,
and the initial valuation $x=y=0$ has bound $[0.6,0.6]$, the correct answer,
which both the SG and backward-reachability methods also return.  No state was
split: the partition of ${\sf z}_0$ by timing choice emerged from the preimage
computation.

\subsection{Minimum Reachability}

For $P_{\min}$ queries, $V_{{\sf z}}$ takes the pointwise \emph{minimum} over
applicable transitions instead of the maximum, and the $\maxintersect$ of
\eqref{eq:and} inverts correspondingly.  The expectation is direction-free, as
are the $[0,0]$ bounds of a symbolic state with no enabled action:
``unreachable under every policy'' holds regardless of the query direction.
Every remaining extremum-selection site branches on the query direction;
Section~\ref{sec:aostar} notes where this matters once the value function is
embedded in a search.

\section{Forward AO* Search}
\label{sec:aostar}

Section~\ref{sec:symba} made each refinement cheap by keeping it in the value
function.  Forward search makes most refinements unnecessary.  \sn employs
And-Or search to explore the symbolic state space and compute bounds on the
probability of satisfying reachability properties, expanding nodes in
best-first order so that refinement effort is spent only where it can still
change the answer at the root.

\subsection{The Search Algorithm}\label{sec:alg}

The two value functions of Section~\ref{sec:symba} correspond to the two node
types of an And-Or graph, which is what lets AO* schedule their computation.
\emph{Or-nodes} represent symbolic states and store $V_{{\sf z}}$; \emph{And-nodes}
represent probabilistic transitions and store $V_\theta$.  Figure~\ref{alg:cap}
gives the control structure, and the paragraphs below refer to its line numbers.

\begin{figure}[htbp]
\footnotesize
\hrule\vspace{2pt}
$\textsc{Symar}({\sf P}, F, \varepsilon)$

\vspace{2pt}\hrule\vspace{3pt}
\begin{tabbing}
\hspace{1.8em}\=xxx\=xxx\=xxx\=\kill
\aln\>$(h, \mathit{Help}) \leftarrow \textsc{SolveSG}({\sf P})$
     \` \emph{// unrefined abstraction}\\
\aln\>$r \leftarrow$ Or-node for ${\sf z}_0$, seeded from $h$\\
\aln\>\textbf{loop}\\
\aln\>\> $N \leftarrow$ unexpanded frontier of the best partial solution from $r$\\
\alnl{l:close}\>\> \textbf{if} $N = \emptyset$ \textbf{then} close the cyclic component below $r$
     \` \emph{// collapse, then relax ungated}\\
\aln\>\> \textbf{else}\\
\alnl{l:help}\>\>\> $B \leftarrow \{u \in N \mid u$ reached by an action in $\mathit{Help}$
      of its parent$\}$, or the best single node if empty\\
\alnl{l:expand}\>\>\> expand each $u \in B$: an And-node per enabled action, an Or-node per outcome
      \` \emph{// growth}\\
\alnl{l:seed}\>\>\> seed each new state: $[1,1]$ if in $F$, $[0,0]$ if no action is enabled,
      else $h$\\
\alnl{l:anc}\>\>\> \textbf{for} each ancestor of $B$ bottom-up over the affected ancestor subgraph\\
\alnl{l:and}\>\>\>\> And-node $\theta = ({\sf z},a,\{{\sf z}_i\})$:
       $V_\theta \leftarrow \maxintersect \{(\pre[]({\sf z},a,\zeta(c)),\
       \textstyle\sum_i \prob(l,a,i)\cdot c_i)\}$
       \` \emph{// refinement}\\
\>\>\>\> \hspace{1.2em} over $c \in \prod_i V_{{\sf z}_i}$ with nonempty preimage,
       or its best two cells plus a hull\\
\alnl{l:or}\>\>\>\> Or-node ${\sf z}$: \ $V_{{\sf z}} \leftarrow
       \maxintersect \{V_\theta \mid \theta \text{ applicable}\}$,
       time-closed if ${\sf z} = r$\\
\alnl{l:term}\>\> \textbf{if} $\textsc{gap}(\textsc{originCell}(V_r)) \leq 2\varepsilon$
     \textbf{then return} $\textsc{originCell}(V_r)$
\end{tabbing}
\vspace{2pt}\hrule
\caption{\label{alg:cap} \sn's control structure.  \textsc{SolveSG} returns admissible
  bounds $h$ and helpful actions from the unrefined stochastic game; the loop
  alternates forward expansion at the frontier with symbolic backpropagation over
  the ancestors, and terminates on the origin cell of the root.}
\end{figure}
\emph{Expansion} (lines~\ref{l:help}--\ref{l:seed}) adds Or-nodes for symbolic states
not previously generated; \emph{backup} (lines~\ref{l:anc}--\ref{l:or}) adds cells to $V_{{\sf z}}$ and $V_\theta$
without splitting the states that hold them or reconstructing their incoming
transitions.  The SG method's refinement does both at once, which is why a split
there propagates structurally; in \sn only expansion adds nodes, and at the
frontier rather than behind it.  A cascade still reaches the ancestors whose bounds
derived from a coarser value, but it revises cells on existing edges.

\emph{Time closure} applies only at the root.  A cell's value is available to any
valuation that can reach it by letting time elapse within the state's zone; for a
non-root state that rewinding is already performed by the incoming transition,
whose $\pret[]$ term maps a successor cell back through post-jump elapse.  The root
has no incoming transition, which is why it needs the step and no other node does.

\emph{Closing} (line~\ref{l:close}) handles a graph in which every node is expanded yet the bounds have
not met.  Members that cannot reach positive-probability mass outside the component
are pinned to $[0,0]$, and the remainder is relaxed with the $\varepsilon$ gate
removed.  The \emph{$\varepsilon$ gate} is the test that lets a node stop being
updated once its interval is narrower than $\varepsilon$, which keeps the
backup from repeatedly re-deriving values that can no longer change the answer.
On a cycle the gate can freeze a loose bound that backup would otherwise tighten,
which is why closing removes it.  Section~\ref{sec:correct} explains what this does
and does not guarantee.

\emph{Termination} (line~\ref{l:term}) reads one cell, not the whole root function.  $V_r$ may hold
several cells, and non-root cells legitimately carry higher values mid-run, so the
maximum over $V_r$ is not the answer to the query.  $\textsc{originCell}(V_r)$
selects the cell whose zone contains the initial valuation, all clocks zero, which
is the only cell the query asks about; $\textsc{gap}$ is the width $ub - lb$ of
that cell.  The interval $[lb,ub]$ is sound, so the true value lies somewhere in it,
and its midpoint is then within $(ub-lb)/2$ of that value.  Comparing the gap
against $2\varepsilon$ therefore certifies a midpoint answer within $\varepsilon$
of the truth, whereas comparing against $\varepsilon$ would demand a midpoint
within $\varepsilon/2$, twice the precision the tolerance asks for.  The
certificate is sound but \emph{a posteriori}: the interval is valid at every
iteration, so an answer returned under the test is always within $\varepsilon$,
but nothing guarantees the gap will reach $2\varepsilon$.  On a cycle a loose
bound may never tighten, in which case the search reports the sound interval
rather than a converged value.  Section~\ref{sec:correct} gives the details.

\subsection{Games Heuristic}\label{sec:heur}

The two methods compose rather than compete.  Before AO* begins, \sn runs the SG
method on the unrefined PTA to compute per-zone bounds $h_{{\sf z}}$, which
initialize the $V_{{\sf z}}$ of open nodes.  One fixpoint solve on the coarsest
abstraction is far cheaper than the refinement cascade it would otherwise drive,
since it is the cascade rather than the initial abstraction that dominates the SG
method's cost.  It is not cheap in absolute terms, however: that abstraction spans
the whole reachable symbolic state space, and on the larger models this solve, not
the search, dominates \sn's runtime, as Section~\ref{sec:results} quantifies.

The SG method is itself an interval method: solving the unrefined abstraction with
both players cooperating gives an upper bound, and with the abstraction player
adversarial a lower bound, so each symbolic state receives a sound interval
$[lb_h,ub_h]$ containing the optimal probability at every valuation it groups.  \sn
seeds an open node only with the side its query optimizes, $ub_h$ for $P_{\max}$ and
$lb_h$ for $P_{\min}$ (the other side stays $0$, respectively the same upper cap).  Cells
bound the value of acting immediately, which delay can only improve for the optimizing
player, so a bound on the optimum is sound for cells on exactly this side
(Section~\ref{sec:correct}); this is why Theorem~\ref{thm:sound} may treat the seed as
sound.  The side must also be admissible or the best-first search order is corrupted.
It is: an unrefined state groups valuations
that may take different optimal actions, so for $P_{\max}$ the SG upper bound
over-approximates the true optimum at every valuation, and dually for $P_{\min}$.
Beyond the bounds, the SG solution also records which actions attain them, which the
next subsection puts to use.
\subsection{Helpful Actions}
\label{sec:helpful}

The SG solve yields more than a bound per state: for each abstract state the choice
set attaining the optimum is also available.  By analogy with heuristic-search
planning \cite{Hoffmann2001FF} we call these the state's \emph{helpful actions}.
\sn uses them as a \emph{focus set} governing how many nodes are expanded per
iteration: instead of expanding one frontier node it expands every frontier node
reached by a helpful action at once, falling back to a single node when none is
helpful.  This lets the relaxation drive the \emph{rate} of expansion rather than
only the order, which matters because backpropagation is expensive relative to
expansion: each round recomputes cross-product preimages along the ancestor path,
so amortizing one backpropagation over several expansions is a real saving.

Unlike the planning technique it is named for, this does not restrict the successor
set: an Or-node always generates a child for every enabled transition, the frontier
is recomputed each iteration, and neither the backed-up values nor the
best-partial-solution walk consult helpfulness.  Admissibility and termination are
therefore independent of the quality of the helpful-action sets, which is what makes
it safe to take guidance from an unrefined abstraction whose choices may be wrong.
\subsection{Best-Only Backup: Partial Refinement}

Forward search decides \emph{which} symbolic states to refine; best-only backup
decides \emph{how far}.  The refinement one And-node induces is the full cell
structure of $V_\theta$, one cell per feasible combination in $C(\theta)$, whose
size grows exponentially in the outcomes and in the cells per successor.
Committing to that partition at every backup is usually wasted, since most
combinations describe sub-regions whose values never reach the root.  Best-only
backup instead commits to the load-bearing sub-regions and summarizes the rest,
exporting at most three cells: the combination $c^*$ with the highest upper bound,
the combination $c^\dagger$ with the highest lower bound if distinct, whose upper
bound is raised to cover any summarized combination overlapping it, and a
\emph{hull} covering the remainder with bounds $[0,\overline{p}^*]$, where
$\overline{p}^*$ is the upper bound of the best summarized combination.  For
$P_{\min}$ every extremum inverts and the hull becomes a pessimistic floor
$[\underline{p}^*,1]$.

The hull is deliberately conservative: its bound is at least as loose as any
summarized combination's true value, and hull cells are barred from being
marked solved, so the deferred part can never masquerade as converged.  Partial
refinement becomes full exactly when the bounds prove to depend on it: if the best
behavior runs through the hull, the And-node is permanently promoted to
enumerating all of $C(\theta)$.  This is the refine-on-demand discipline of the
search as a whole, applied within one backup.

\section{Correctness}
\label{sec:correct}

Write $P^*({\sf z},v)$ for the true optimal probability of reaching $F$ from the
concrete state $(l,v)$ of ${\sf z}=(l,\zeta)$, the maximum over policies for a
$P_{\max}$ query and the minimum for a $P_{\min}$ one.  The quantity the query asks
for is $P^*$ at the initial state: $p_{\sf P}^{\mathit{opt}}(F)$ denotes
$p_{\sf P}^{max}(F)$ or $p_{\sf P}^{min}(F)$ of Definition~\ref{def:reach}
according to that direction, so $p_{\sf P}^{\mathit{opt}}(F) = P^*({\sf z}_0,
\mathbf{0})$ with $\mathbf{0}$ the all-zero valuation.

The backup works with a second, narrower quantity.  Write $P^0({\sf z},v)$ for the
optimal probability when an action is taken at $v$ \emph{immediately}, with no
delay first: $1$ if $l\in F$, and $0$ if no action is enabled at $v$.  Delay is
the only difference, so
\[
P^*({\sf z},v) \;=\; \mathrm{opt}_{d\ge 0,\; v+d\in\zeta}\; P^0({\sf z},v+d),
\]
with $\mathrm{opt}$ the supremum for $P_{\max}$ and the infimum for $P_{\min}$.
The two differ whenever delaying enables an action: a valuation at which nothing
is enabled has $P^0=0$ but may have $P^*>0$.

\begin{definition}[Soundness invariant]\label{inv}
$V_{{\sf z}}$ is \emph{sound} if for every cell $(\zeta_i,[lb_i,ub_i]) \in
V_{{\sf z}}$ and every $v \in \zeta_i$, \ $lb_i \le P^0({\sf z},v) \le ub_i$.
The \emph{time-closed} value function $\overline{V}_{{\sf z}}$ obtained by the
closure of Section~\ref{sec:alg} is sound if the same holds with $P^*$ in place of
$P^0$.
\end{definition}

The invariant is cell-wise and says nothing about valuations in no cell: the merge
resolves overlapping cells without guaranteeing that their union covers $\zeta$, so
coverage matters only at the root, whose origin cell must exist for the reported
answer to mean anything.  Cells below the root bound $P^0$, not $P^*$; delay
before an action enters through $\pret[]$ on the successor side of each backup
(Lemma~\ref{lem:pre}) and, for the root, through the closure.

\begin{lemma}[Preimage]\label{lem:pre}
Let $c$ select cell $(\zeta_i,\cdot)$ from each successor ${\sf z}_i$.  Then
$\pre[]({\sf z},a,\zeta(c))$ is the set of $v \in \zeta$ at which $a$ is enabled and
from which executing $a$ immediately can produce, for every outcome $i$, an
arrival valuation $w_i$ that can elapse time into $\zeta_i$.
\end{lemma}

What must be common is the \emph{source valuation} at which $a$ is taken, not the
behavior after it: nature does not reveal the outcome in advance, so a single $v$
must serve every outcome.  The post-jump trajectories may differ, and
Definition~\ref{def:pre} allows that, applying $\pret[]$ per outcome inside the
conjunction.  Each successor zone is closed under post-jump elapse, so $\pret[]$
carries a successor cell back to that outcome's jump instant; $\prej[]$ then inverts that
outcome's resets, and the results are intersected in the source frame, where they
constrain the one valuation at which $a$ occurs.

\begin{theorem}[Soundness]\label{thm:sound}
Every $V_{{\sf z}}$ and $V_\theta$ \sn produces is sound in the sense of Definition~\ref{inv}.  In particular, if the origin-containing
root cell is $[lb,ub]$ then $lb \le p_{\sf P}^{\mathit{opt}}(F) \le ub$.
\end{theorem}
\begin{proof}[Sketch]
Induction over the backup sequence, for $P^0$ below the root.  \emph{Initialization.}
A state with location in $F$ has $P^0=1$ and a valuation at which no action is
enabled has $P^0=0$, so $[1,1]$ and $[0,0]$ are exact and direction-independent
(these cells say nothing about $P^*$ at such a valuation, which time may yet
carry into a guard); otherwise the seed uses the side of $h$ that the query
optimizes, which bounds $P^0$ because it bounds $P^*$ in the direction where
$P^0$ lies inside it (Section~\ref{sec:heur}), and the other side is $0$ for
$P_{\max}$ or the same upper cap for $P_{\min}$; $[0,1]$ absent a heuristic.
\emph{Transition backup.}  Fix feasible $c$ and $v \in \pre[]({\sf z},a,\zeta(c))$.
By Lemma~\ref{lem:pre}, taking $a$ at $v$ gives each outcome $i$ an arrival $w_i$
that can elapse into $\zeta_i$ at some $w_i' \ge w_i$.  A policy may do exactly
this, so $P^*(w_i) \ge P^0(w_i') \ge lb_i$ for $P_{\max}$ (dually $\le ub_i$ for
$P_{\min}$).  Define the \emph{value through cell} $\zeta_i$ as $P^0(w_i')$
bracketed by $[lb_i,ub_i]$; since $\prob(l,a,i) \ge 0$ and sums to one,
monotonicity of expectation gives $\sum_i \prob(l,a,i)\,lb_i \le P_c(v) \le
\sum_i \prob(l,a,i)\,ub_i$ for the value $P_c(v)$ of taking $a$ at $v$ and then
delaying into the cells selected by $c$.  This is the interval assigned.  The
optimum over delays is then the optimum over $c$: each outcome elapses
independently, so the cross product $C(\theta)$ ranges over every choice of
landing cell, and $P^0_\theta(v)=\mathrm{opt}_c P_c(v)$ over the combinations
$c$ with $v\in\pre[]({\sf z},a,\zeta(c))$.  Where preimages overlap,
$\maxintersect$ retains one whole candidate rather than combining bounds: for
$P_{\max}$ it keeps the candidate $j$ of greatest $ub$, breaking ties on $lb$.
This is sound.  Each candidate brackets its own value, $lb_i \le P_i \le ub_i$,
so $\max_i P_i \le \max_i ub_i = ub_j$; and $lb_j \le P_j \le \max_i P_i$, so
$[lb_j,ub_j]$ contains $\max_i P_i$.  The upper bound needs every landing cell to
be a candidate, so it requires that the successor cells cover every valuation at
which some action is enabled; the remainder carries $[0,0]$.  For $P_{\min}$ the
dual argument applies to the candidate of least $lb$.  \emph{State backup.}  The
same argument applies to the extremum over applicable transitions, the Bellman
choice among actions enabled at $v$; it bounds $P^0$.  \emph{Root closure.}  The
closure rewinds each root cell through time within $\zeta$, best cell first, so a
valuation $v'$ that can elapse to $v$ inherits the cell's bound.  For $P_{\max}$
the lower bound is valid because $P^*(v') \ge P^0(v)$.  The upper bound is valid
because the cells are processed in descending $ub$: $v'$ is claimed by the
highest-$ub$ cell that rewinds onto it, which bounds $\sup_d P^0(v'+d)=P^*(v')$.
\emph{Partial refinement.}  The retained $c^*$ carries the interval full backup
assigns it.  The retained $c^\dagger$ carries the interval full backup assigns
it, except that where a dropped combination overlaps $c^\dagger$ outside $c^*$ the
cell takes the weaker of the two bounds on that region (the larger $ub$ for
$P_{\max}$), since the dropped combination's candidate would otherwise have won
there under $\maxintersect$.  The hull is at least as loose as every combination
it summarizes, so it cannot exclude $P^0$; its lower bound is $0$ because it may
cover valuations no combination realizes.  Promotion to full enumeration only
narrows it.  Soundness is independent of convergence: it holds at every
iteration, so the bounds may be reported whether or not the tolerance was met.
\end{proof}
\paragraph{Convergence.}
Soundness does not imply the bounds meet.  On a finite acyclic solution graph that is
fully expanded and whose load-bearing deferred backups have all been promoted, they
do, by induction on distance from the exact target and dead-end nodes.  Both provisos
matter: an unexpanded leaf still carries its heuristic interval and an unpromoted hull
a placeholder lower bound.  Cycles are the further difficulty, since a node on a cycle
can hold a loose bound that backup never tightens, its value depending on its own
previous value.  The closing phase of Section~\ref{sec:alg} attacks this in the two
ways described there, neither guaranteed to close every gap.  \sn therefore treats
convergence as something to check rather than assume: it accepts an answer only
when the origin cell's gap is within tolerance, and if relaxation stalls it reports
the sound interval it has instead of claiming a converged value.
Section~\ref{sec:results} shows the bounds do close, to tolerance, on every
benchmark.  A general guarantee is future work; the missing ingredient is a
treatment of cycles a policy can stay inside forever, and it must operate on
(node, cell) pairs rather than nodes, since a symbolic state may cycle for some
clock valuations and escape for others.

\section{Related Work}\label{sec:related}

Kwiatkowska et al.\ \cite{KNP09c} established the foundation for PTA verification,
formulating reachability as a stochastic game and obtaining two-sided bounds by
abstraction refinement.  \sn is a direct descendant: we retain the central idea
that sound bounds come from refining an abstraction until the property-relevant
zone distinctions are exposed, and change only how refinement is represented and
scheduled, as Section~\ref{sec:mdp} describes.  Because their method remains the
best available source of admissible bounds, \sn uses it as its heuristic rather
than discarding it.  PTA model checking otherwise divides into backward
reachability \cite{kwiatkowska2007symbolic} and forward search
\cite{kwiatkowska2002automatic}; \sn is a forward method that avoids the blowup
usually afflicting them by refining symbolically rather than enumerating.  We
previously developed an AO* MDP solver PRISM extension~\cite{goldman:difts15}, and Hansen
and Zilberstein's LAO* \cite{HANSEN200135} handles the cycles plain AO*
cannot; the closing phase of Section~\ref{sec:alg} plays that role here.

Symbolic value iteration \cite{Hoey1999Apricodd,StAubin2000SPUDD} represents MDP
value functions as algebraic decision diagrams.  Like \sn these avoid enumerating
states, but their basis is a fixed propositional one and cannot express the
continuous zone partitions timed transitions induce, whereas \sn's partition is
built on demand from preimage computation.  Point-based POMDP methods
\cite{Spaan2005Perseus,Smith2004HSVI,Kurniawati2008SARSOP} are closer still:
HSVI in particular tightens two-sided bounds by heuristically guided forward
search, and its belief simplex plays the role our clock zone does.  The difference
is the representation's shape: $\alpha$-vectors are globally linear and their
hull convex, whereas \sn's zone partition is piecewise-constant and non-convex,
arising from the geometry of clock resets and guards.  Statistical model checking
\cite{6354636,BulychevUPPAALSMCStatisticalModel2012} samples instead of
enumerating, which suits finding counterexamples to a bound but not proving that
no property-violating policy exists.

\section{Experiments and Results}
\label{sec:results}
We implemented \sn within PRISM~\cite{Kwiatkowska2011PRISM4} and compare it with the SG method, the
state-of-the-art baseline, on the PRISM benchmark PTAs.  \sn runs with the games
heuristic, helpful actions, and best-only backup all enabled (Section~\ref{sec:aostar}).
Table~\ref{tab:results} compares the two methods on both $P_{\max}$ and $P_{\min}$
instances; PRISM's backward reachability engine does not implement $P_{\min}$, so we
report it in the text rather than the table.

\subsection{PTA Instances}

We used the PTA models distributed with PRISM\footnote{\href{https://www.prismmodelchecker.org/casestudies/}{https://www.prismmodelchecker.org/casestudies/}}:
the non-repudiation security protocol with an honest principal ($P_{\min}$) and with
a malicious one ($P_{\max}$); IEEE 1394 FireWire leader election in both its abstract
and implementation models ($P_{\min}$); the CSMA/CD protocol in its abstract
($P_{\min}$) and full ($P_{\max}$) models; and the Zeroconf address assignment
protocol ($P_{\max}$).  Except for csma/full, parameterized by its collision bound
$K$, each is queried under deadlines at a range of constants $T$; the deadline
queries are where the methods differ most.  On the small formats09 example of
Figure~\ref{fig:ex-pta} all three methods are similar, and we omit it.

\subsection{Experimental Setup}

All experiments ran on an Apple M4 Max, in an amd64 Linux container (Alpine Linux~3.24)
allotted $10$ CPUs and $96$\,GB of memory.  \sn and the SG method are both implemented in
PRISM~4.8.1 and ran on OpenJDK~21.0.12 with a $2$\,GB maximum heap.  Each instance was run
three times and we report the mean of the time PRISM reports as ``model checking
completed'', which covers constructing the PTA and the check, including for \sn the
heuristic solve, but not JVM startup or parsing; a run stopped at the $1200$\,s limit or
by heap exhaustion was not repeated and appears as a dash in Table~\ref{tab:results}.
Games used each benchmark's distributed refinement options (\texttt{refine=all}).

\subsection{Results}

Each method states its stopping condition differently: games refines until the
initial-state gap is below its refinement parameter, BR runs value iteration to the same
absolute criterion, and \sn accepts an origin cell whose gap is at most $2\varepsilon$.
We set $\varepsilon = 5\times10^{-7}$ for \sn (window $10^{-6}$) and the refinement
parameter to $10^{-6}$ for games; results agree to within $10^{-6}$ on every instance
both finished (largest difference $5.4\times10^{-7}$), including BR on the $P_{\max}$
instances, an independent method rather than a refinement of the same abstraction.
\begin{table}[htbp]
\centering
\footnotesize
\setlength{\tabcolsep}{4pt}
\begin{tabular}{|l|r r r r|r r r r|r|}
  \hline
  & \multicolumn{4}{c|}{Games} & \multicolumn{4}{c|}{SYMAR} & \\
  Instance / Property & init & final & ref & s
    & gen & sol & iter & s & $P(\text{prop})$ \\
  \hline\hline
  \multicolumn{10}{|l|}{\emph{repudiation/honest ($P_{\min}$)}} \\
  \hline
  \quad deadline $T=40$  & 171  & 471   & 19  & 0.31
                         & 8 & 8 & 6 & \textbf{0.25} & 0.61258 \\
  \quad deadline $T=400$ & 1611 & 31611 & 199 & 25.70
                         & 8 & 8 & 6 & \textbf{0.50} & $\approx 1.0$ \\
  \quad deadline $T=524$ & 2107 & 53221 & 242 & 55.35
                         & 8 & 8 & 6 & \textbf{0.51} & $\approx 1.0$ \\
  \quad deadline $T=526$ & 2113 & 2113  & 0   & \textbf{0.42}
                         & \multicolumn{3}{c}{seed} & 0.50 & $\approx 1.0$ \\
  \hline
  \multicolumn{10}{|l|}{\emph{firewire/abst ($P_{\min}$, delay=360)}} \\
  \hline
  \quad deadline $T=10000$ & 525  & 1020  & 19 & 0.55
                           & 23 & 19 & 12 & \textbf{0.33} & 0.97473 \\
  \quad deadline $T=20000$ & 1875 & 9070  & 40 & 3.74
                           & 23 & 19 & 12 & \textbf{0.44} & 0.99963 \\
  \quad deadline $T=33000$ & 4850 & 43678 & 44 & 45.62
                           & 23 & 19 & 12 & \textbf{0.64} & $\approx 1.0$ \\
  \quad deadline $T=34000$ & 5141 & 6001  & 1  & 1.13
                           & 23 & 19 & 12 & \textbf{0.69} & $\approx 1.0$ \\
  \hline
  \multicolumn{10}{|l|}{\emph{firewire/impl ($P_{\min}$, delay=360)}} \\
  \hline
  \quad deadline $T=5000$  & 3847  & 4463  & 17 & 1.37
                           & 37 & 33 & 20 & \textbf{0.79} & 0.78125 \\
  \quad deadline $T=10000$ & 12168 & 24449 & 56 & 18.74
                           & 37 & 33 & 20 & \textbf{1.46} & 0.97473 \\
  \quad deadline $T=15000$ & 26171 & 86380 & 73 & 234.70
                           & 37 & 33 & 20 & \textbf{2.43} & 0.99719 \\
  \hline
  \multicolumn{10}{|l|}{\emph{csma/abst ($P_{\min}$, bmax=1, K=4)}} \\
  \hline
  \quad deadline $T=1000$ & 80598  & 80598  & 0  & \textbf{2.32}
                          & \multicolumn{3}{c}{unreach} & 2.63 & 0.0 \\
  \quad deadline $T=2000$ & 353715 & 359811 & 16 & 31.08
                          & 32 & 28 & 23 & \textbf{12.18} & 0.40537 \\
  \quad deadline $T=2500$ & 582518 & 695906 & 17 & 870.73
                          & 32 & 28 & 23 & \textbf{18.81} & 0.94908 \\
  \quad deadline $T=3000$ & \multicolumn{3}{c}{---} & ---
                          & 32 & 28 & 23 & \textbf{29.65} & 0.99642 \\
  \hline\hline
  \multicolumn{10}{|l|}{\emph{repudiation/malicious ($P_{\max}$)}} \\
  \hline
  \quad deadline $T=5$  & 1660  & 1700  & 2  & \textbf{0.34}
                        & 26   & 10  & 8   & 0.38 & 0.1 \\
  \quad deadline $T=10$ & 5641  & 6589  & 15 & 1.36
                        & 470  & 197 & 218 & \textbf{0.95} & 0.10544 \\
  \quad deadline $T=40$ & 159272 & 172687 & 6 & 53.07
                        & 8806 & 4897 & 4391 & \textbf{6.49} & 0.10566 \\
  \hline
  \multicolumn{10}{|l|}{\emph{zeroconf ($P_{\max}$)}} \\
  \hline
  \quad deadline $T=200$ & 811  & 858  & 16 & \textbf{0.31}
                         & 407  & 399  & 340  & 0.73 & 0.00122 \\
  \quad deadline $T=400$ & 3264 & 6776 & 31 & 2.03
                         & 3234 & 3179 & 2880 & \textbf{1.67} & 0.00130 \\
  \quad deadline $T=2000$ & 66011 & 66011 & 0 & \textbf{2.61}
                          & \multicolumn{3}{c}{seed} & 2.90 & 0.00130 \\
  \hline
  \multicolumn{10}{|l|}{\emph{csma/full ($P_{\max}$, bmax=2, COL=4)}} \\
  \hline
  \quad collisions $K=2$ & 6385  & 6487  & 10 & \textbf{0.88}
                         & 364 & 325 & 292 & 1.19 & 0.14355 \\
  \quad collisions $K=4$ & 34715 & 34849 & 10 & \textbf{2.07}
                         & 644 & 597 & 444 & 2.27 & 0.07690 \\
  \quad collisions $K=8$ & 818139 & 818273 & 10 & \textbf{24.61}
                         & 644 & 597 & 444 & 26.62 & 0.07690 \\
  \hline
\end{tabular}
\caption{\label{tab:results} Runtime and search-space size, faster time in bold.
  For games: initial and final STPG states, refinement rounds, and seconds.  For
  \sn: nodes generated, final solution-graph size, AO* iterations, and seconds.
  $P_{\min}$ families above the double rule, $P_{\max}$ below.  A dash marks a
  method that exceeded the $1200$\,s limit or its $2$\,GB heap.  ``seed'' marks instances whose
  unrefined heuristic already met the acceptance width, so AO* did not search;
  ``unreach'' those it settled by proving every upper bound zero.}
\end{table}

\paragraph{Runtime.}
On the $P_{\min}$ families \sn is faster by one to two orders of magnitude at
intermediate deadlines ($108\times$ on repudiation/honest at $T=524$, $97\times$ on
firewire/impl at $T=15000$), and on csma/abst
by $2.6\times$ at $T=2000$ and $46\times$ at $T=2500$ ; at
$T=3000$ it returns in $29.7$\,s where games exhausts the $1200$\,s limit.  Games leads only where refinement is cheap
(the bold games entries of Table~\ref{tab:results}), never by more than about $2.0$\,s, and
\sn is at most $35\%$ slower on all of them but zeroconf at $T=200$ ($0.31$\,s against
$0.73$\,s).

The two degrade differently.  Games' cost tracks refinement depth, which is non-monotone
in $T$: $55.4$\,s at $T=524$ on repudiation/honest against $0.42$\,s at $T=526$, as the
refinement count falls from $242$ to $0$.  Its state count grows with the deadline,
whereas \sn's search graph does not ($37$ generated nodes and $20$ iterations on every
firewire/impl instance, $8$ and $6$ on repudiation/honest, $23$ and $12$ on
firewire/abst), because \sn expands only along the best partial solution.  The rows \sn
loses are not lost in the search, which is always negligible (csma/full $K=8$: $0.53$\,s
of $26.6$\,s): the cost is solving the unrefined abstraction over the entire reachable
symbolic state space ($818{,}139$ states there), so \sn cannot beat games by more than
games spends on \emph{refinement} beyond that solve.  Wins track refinement rounds, and
csma/full, needing $10$ rounds at every size, admits none.

PRISM's backward reachability engine (only the $P_{\max}$ instances) is fastest on
csma/full and on repudiation/malicious at $T=10$, but degrades with the state space it enumerates: on zeroconf at $T=400$ it needs
$102$ value-iteration rounds over a $3165$-state backward graph, taking $17.6$\,s against
\sn's $1.7$\,s.

\section{Conclusion}
\label{sec:conclusion}

We have presented \sn, which adapts the abstraction refinement of Kwiatkowska et
al.~\cite{KNP09c} along two axes: refinement is \emph{symbolic}, the partition the SG
method materializes as split states appearing instead in a zone-valued probability
function $V : \Zones \rightarrow [0,1]^2$ updated in place, and \emph{demand-driven},
directed by forward AO* search with bounds and helpful actions from the unrefined SG
abstraction and partial refinement bounding the cross product.  The storedabstraction and partial refinement bounding the cross product.  The stored
intervals are sound at every iteration independently of convergence
(Theorem~\ref{thm:sound}), so a run stopped early still bounds the answer.

The method pays off where refinement is expensive and the property-relevant fragment is small: \sn is up to $108\times$ faster than the SG method, solves a csma instance
on which that method exhausts the time limit, and holds its search graph constant as
the deadline grows where the SG abstraction expands with it.  Where refinement is
cheap the two are within about two seconds of each other, because \sn's cost is dominated by the
heuristic solve ($98\%$ of the total on the largest csma/full instance).  Lowering that
floor remains open, as does convergence, which we certify only \emph{a posteriori}.

\paragraph{AI Usage.}

This paper's prose was drafted with the assistance of a large language model (Claude,
Anthropic) under author direction; the authors specified all content and claims and
revised all generated text.  The same assistant scripted the benchmark runs and
instrumented the implementation for Table~\ref{tab:results}.  The authors reviewed
the benchmark scripts and the implementation.  No results or citations
were model-generated, and the algorithm, correctness argument, and implementation are
the authors' own.

\paragraph{Acknowledgements} This material is based upon work supported by the
Defense Advanced Research Projects Agency under Contract No. HR00112490439. Any
opinions, findings and conclusions or recommendations expressed in this material
are those of the author(s) and do not necessarily reflect the views of the
Defense Advanced Research Projects Agency.

\bibliographystyle{eptcs}
\bibliography{main}

\end{document}